\documentclass[]{ceurart}

\usepackage{hyperref}
\usepackage[all]{hypcap}
\usepackage{listings}

\begin{document}

\copyrightyear{2024}
\copyrightclause{Copyright for this paper by its authors.\\
  Use permitted under Creative Commons License Attribution 4.0 International (CC BY 4.0).}

\conference{IVUS2024: Information Society and University Studies 2024, May 17, Kaunas, Lithuania}

\title{Inference acceleration for large language models using ``stairs'' assisted greedy generation}

\author[1]{Domas Grigaliūnas}[%
orcid=0009-0006-8664-0544,
email=domas.grigaliunas@ktu.edu,
url=https://domas.info/
]
\cormark[1]
\fnmark[1]
\address[1]{Kaunas University of Technology,
  K. Donelaičio g. 73, Kaunas, 44249, Lithuania }

\author[1]{Mantas Lukoševičius}[%
orcid=0000-0001-7963-285X,
email=mantas.lukosevicius@ktu.lt,
url=https://mantas.info/
]
\fnmark[1]

\begin{abstract}
Large Language Models (LLMs) with billions of parameters are known for their impressive predicting capabilities but require lots of resources to run. With their massive rise in popularity, even a small reduction in required resources could have an impact on environment. On the other hand, smaller models require fewer resources but may sacrifice accuracy. In this work, we are proposing an implementation of ``stairs'' assisted greedy generation. It is a modified assisted generation methodology that makes use of a smaller model's fast generation, large model's batch prediction, and ``stairs'' validation in order to achieve a speed up in prediction generation. Results show between 9.58 and 17.24 percent inference time reduction compared to a stand-alone large LLM prediction in a text generation task without a loss in accuracy.
\end{abstract}


\maketitle

\section{Introduction}

Large Language Models (LLMs) with billions of parameters are known for their impressive predicting capabilities but require lots of resources (hardware, computation time, energy) to run. With their rapid rise in popularity, this is also becoming an environmental issue, among others. Even a small reduction in required resources can have a big global impact. On the other hand, smaller models require fewer resources but may sacrifice accuracy. 

In this work, we explore a solution that combines the strengths of both a large and a small language models, aiming to have faster inference without reducing the accuracy of prediction. We propose a novel code implementation of a methodology for inference time reduction. The idea is that the smaller model generates several tokens in advance and ``stairs'' batch validation detects how many next token predictions can the main LMM skip. It exploits the fact that an LMM can generate several similar independent next token predictions (a batch) in a single iteration in parallel with relatively small computational overhead compared to a single prediction. This saves expensive iterations for the main model in exchange for several significantly cheaper predictions from a smaller model. We are calling it a ``stairs'' assisted greedy generation. Results indicate between 9.58 and 17.24 percent inference time improvement for text generation without sacrificing accuracy compared to text generation by a single LLM itself.

In Section \ref{review}, we explain inspiration, related works, and models fused for our experimentation. Section \ref{methods} contains explanations about LLMs' next token prediction towards workings of ``stars'' assisted greedy generation. Section \ref{results} contains all the relevant information about the experiments and results. Finally, Section \ref{conclusions} summarises conclusions and provides directions for future works.

\section{Literature review}\label{review}

This section contains information about the inspiration for the experiments, related research, and models used in experimentation.

\subsection{Inspiration}

The idea for experimentation was greatly inspired by a tweet written by Andrej Karpathy \cite{tweet}. The main ideas are that batch prediction has a similar or marginally higher retrieval cost compared to a single prediction, large models have a memory bottleneck when predicting, and, assuming we have some computing power remaining, assistant models could generate extra prompts to use with batch prediction. 

\subsection{Related works}

Recent research has explored various approaches to accelerate inference in large language models. Notably, Google \cite{leviathan2023fast} proposed speculative decoding, a technique that utilizes a combination of smaller models and rejection sampling, a technique that picks points from an easy-to-sample distribution, removes them if they do not fit the target distribution, and repeats until it gets enough good ones \cite{clifford1994monte}, to generate multiple candidate tokens in parallel from a large autoregressive model. Specifically, this approach allowed the T5-large model (770 million parameters) \cite{T5} to achieve between 1.4 and 1.7 times speedup in English to German translation task of WMT 2018 \cite{leviathan2023fast} returning identical outputs.

Additionally, DeepMind \cite{chen2023accelerating} introduced speculative sampling with a focus on leveraging small models to parallelize token generation from a larger target LLM. Their method employs a novel rejection sampling scheme to ensure the generated text adheres to the highest next token in a probability distribution of the sequence. For Chinchilla (70 billion parameters) model \cite{hoffmann2022training} with an undisclosed 4 billion parameters assistant model, the proposed methodology achieved between 1.92 and 2.04 times of speedup for The Extreme Summarization (XSum) \cite{narayan1808don} task while empirically verifying that outputs come from the same distributions for regular and speculative sampling inferences. 

In contrast to these speculative decoding and sampling methods, our initial focus was only to explore a greedy generation solution. At the end of the works, the closest available implementation was identified from HuggingFace \cite{gante2023assisted} research in their Transformers library-assisted generation. A flan-t5-large (780 million parameters) model, which, in short, is a fine-tuned t5-large version \cite{chung2024scaling} with flan-t5-small (60 million parameters) assistant model using a proposed methodology achieved 25.91 percent speedup for ``CNN Dailymail'' task of articles summarisation  \cite{chen2016thorough}. Additionally, for a flan-t5-xl (3 billion parameters) model with the same, flan-t5-small model assistant, assisted generation achieved 26 percent inference improvement for the same task. Final accuracy was not described.

For our work, the accuracy of models was tested with BLEU (BiLingual Evaluation Understudy) score \cite{BLEU} in the range from 0 to 100, where 75-100 stands for a perfect score, with 100 being an identical match.

\subsection{Models}

In experiments we selected a group of autoregressive T5 (Text-to-Text Transfer Transformer) models \cite{T5}:
\begin{itemize}
    \item T5-small -- 60 million parameter version. Used as an assistant model.
    \item T5-large -- 770 million parameters. Used as the main model for T5-large experiments.
    \item T5-3B -- 3 billion parameters. Used as the main model for T5-3B experiments.
\end{itemize}

The main reason for this selection was to recreate a similar environment where previous experiments -- Google's speculative decoding with T5-large and HuggingFace's assisted generation with flan-t5-large and xl -- were run. Additionally, DeepMind's Chinchilla model and its quantized and pruned versions were not considered due to the huge size and unknown assistant model. 

\section {Methodology}\label{methods}

This section explains the next token prediction using LLMs, a methodology of assisted generation, and the proposed ``stairs'' assisted greedy generation.

\subsection{Large language model next token prediction}

\begin{figure}[hbt!]
\centering
\includegraphics[width=0.55\textwidth]{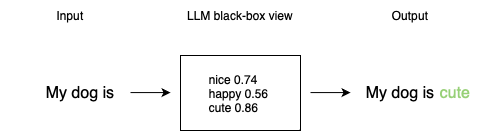}
\caption{An abstract example of the LLM next token prediction. The token ``cute'' is chosen because it has the highest probability in the vocabulary. The probability is calculated based on a given input. }
\label{nextToken}
\end{figure}

In abstract terms, a large language model uses given tokens as input and estimates the next best token from the vocabulary using its learned knowledge. For example, given the text ``My dog is'' as an output we might receive ``My dog is cute'' as a result because the token ``cute'' will have the largest probability to be chosen. This is illustrated in Figure \ref{nextToken}. 

\begin{figure}[hbt!]
\centering
\includegraphics[width=0.6\textwidth]{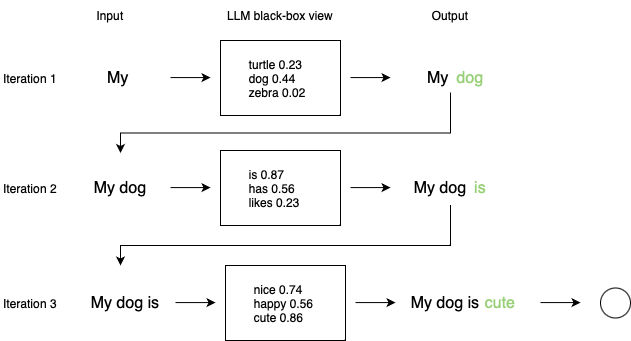}
\caption{An abstract example of the LLM next several tokens prediction. Specifically, showing how the model greedily chooses the next token in each iteration. This example took 3 iterations.}
\label{figLMM} 
\end{figure}

Essentially, by repeating this step the model predicts a full sequence and returns the final output. An example can be seen in Figure \ref{figLMM}. So if the model is large and relatively slow, generating each token takes a noticeable amount of time and resources. This is where assisted generation comes to help.

\subsection {Assisted generation}

Assisted generation is based on a speculative execution optimization technique which in short could be described as a methodology where the processor performs several tasks in advance in order to have the results faster when such are needed \cite{speculative}.

\begin{figure}[hbt!]
\centering
\includegraphics[width=0.6\textwidth]{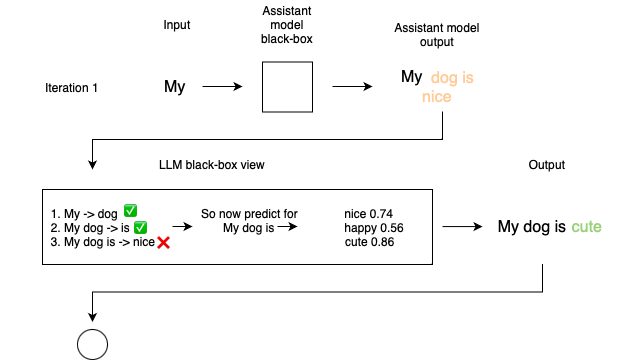} 
\caption{An abstract example of the LLM next several tokens prediction with an assistant model emphasizing how the assistant model generates several tokens in advance. The large model keeps only those tokens that are greedily matched, and using them generates the following tokens. This example took one large model iteration.}
\label{figAS}
\end{figure}

In the assisted generation case, a smaller -- assistant -- model performs predictions for one or several next tokens and gives generated prompts as input for the main model. Such prompts are evaluated by checking them one by one, from left to right while only keeping the ones that match greedily with their highest probability in the vocabulary. This step is repeated adding one more token each time until they run out or mismatches are found. After such input is ready, the model as usual predicts the next token. Such a cycle repeats until the prompt is fully responded to. 

For performance gains, if any of the smaller model predicted tokens are accepted this removes one loop of iteration from the main model. This is illustrated in Figure \ref{figAS}.

\subsection {``Stairs'' assisted greedy generation}

\begin{figure}[hbt!]
\centering
\includegraphics[width=0.6\textwidth]{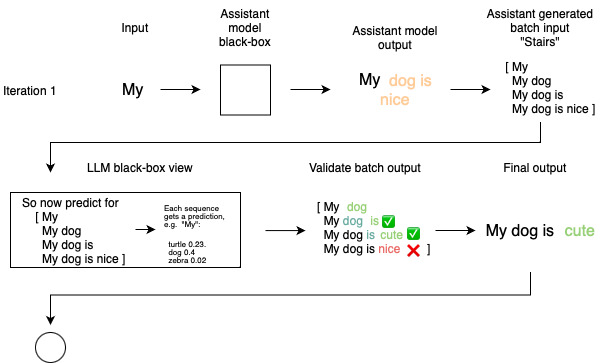}
\label{fig:figstairspred}
\caption{An abstract example of the LLM next several tokens prediction with ``stairs'' batch prediction of an assistant model emphasizing how the assistant model generates several tokens in advance and puts them into the batch. The large model predicts each of them individually. After prediction, ``stairs'' greedy selection is applied and only validated tokens are kept. This example took one iteration of the large model to generate the final result.}
\end{figure}

In ``stairs'' assisted generation, the initial steps for an assistant model are identical -- a smaller model generates a prompt. Next, the initial sentence is broken down into subsequences that start with the first word and progressively include one more token until the full prompt is reached. All of them are combined in the same order as the batch. The matrix is given to the main model and then the generated output contains one additional token for each sequence. A visual representation can be seen in Figure \ref{fig:figstairspred}.

After the output is received, ``stairs'' batch validation is performed. The ``stairs'' batch validation is inspired by incremental validation and teacher forcing algorithm. The algorithm trains recurrent networks by providing observed values as inputs during training \cite{teacherforcing}. In our case, instead of training we perform validation with a matrix of sequences and are also providing already predicted values. The testing design of incremental validity \cite{incrementalvalidity}, in our case, is applied for every comparison between two sequences. Only if they match, excluding a last token of a second sequence, we update the ground truth with the last token of a second sequence.

\begin{figure}[hbt!]
\centering
\includegraphics[width=0.75\textwidth]{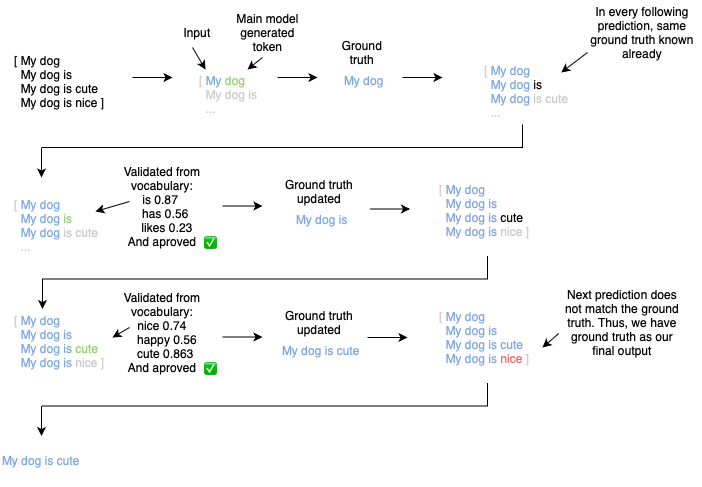}
\label{fig:figstairs}
\caption{ ``Stairs'' greedy validation example that is applied after the main model predicts a given batch. The first vector is taken as a ground truth, at the same time approving tokens vertically below. Next vector's final token is checked against the vocabulary and if approved new ground truth is set. The process continues until the end of a batch or when the ground truth does not match the next vector prediction. The final ground truth is returned as a final output for a given prediction iteration. }
\end{figure}

The ``stairs'' batch validation begins with the very first vector being set as the ground truth. The following vector is validated using the current ground truth to check the vocabulary. If the check is successful, a new ground truth is set. The process continues until the model checks each sequence or ground truth does not match the next sequence. The ground truth is returned as an output. All this sequence is illustrated in Figure \ref{fig:figstairs}.

The main benefit of such methodology is that in cases where inference limitations come from the slowness of information retrieval, for a marginal latency increase a significantly increased throughput can be obtained \cite{gante2023assisted}. A prerequisite for such a trade-off is to have similar prompts in the batch. The general idea why this works is that cached model weights are reused for the similar inputs instead of reloading weights each time \cite{offloading}.

\section {Experiments and results}\label{results}

This section contains information about a general setup, the two main experiments: T5-large and T5-3B, and their results.

\subsection {Experiments setup}

Both experiments had a single main model: either T5-large or T5-3B. Each experiment had two stages:
\begin{enumerate}
    \item ``Stairs'' assisted greedy generation best batch size analysis.  
    \item Main model versus HuggingFace assisted generation versus ``Stairs'' assisted greedy generation model text prediction.
\end{enumerate}

HuggingFace did not have a batch size analysis step since it sets it dynamically.

Testing environment: MacBook PRO M2, 32GB RAM.
Functional code is implemented reusing HuggingFace transformers library \cite{hftransformer} and modifying Greedy generation code path.

\subsection {Input data}

Each experiment was tested with a single prompt: ``translate English to German: My dog is cute.''. The main reasons for this choice was to have a faster start testing implementation and it was one of the most used forms of example prompts in HuggingFace.co.

\subsection{T5-large}

\begin{figure}[hbt!]
\centering
\includegraphics[width=0.5\textwidth]{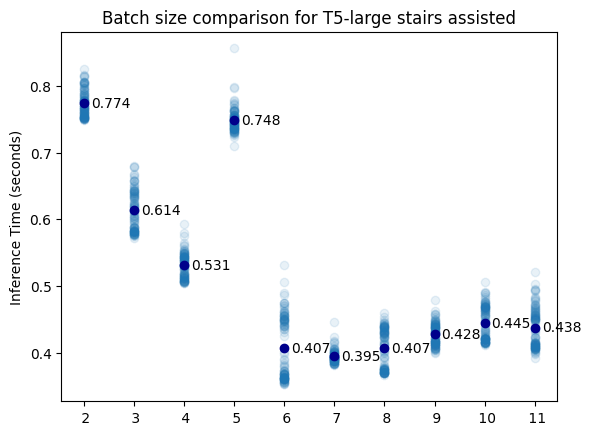}
\label{fig:fig1}
\caption{ T5-large stairs assisted generation model inference time in seconds based on batch size (length of assistant model prediction) comparison. The figure displays the distribution of resulting inference times for 100 independent generations of each batch size (1000 runs in total).} 
\end{figure}

T5-large with an assistant of T5-small-stairs was tested for the best batch size, checking each one 100 times after a warmup (one run) against the prompt. The investigation started from batch size 2 since 1 means predicting the same way as the large model. Results indicate that batch size has a noticeable effect on model performance, with the best case being batch size 7. All generated responses score between 75 and 100 in BLEU score. Figure \ref{fig:fig1} illustrates the results.

After that, three cases were compared:
\begin{itemize}
    \item Original -- T5-large
    \item HF assisted -- HuggingFace T5-large with assistance of T5-small
    \item Stairs assisted -- T5-large with assistance of T5-small-stairs with batch size 7.
\end{itemize}

\begin{figure}[hbt!]
\centering
\includegraphics[width=0.5\textwidth]{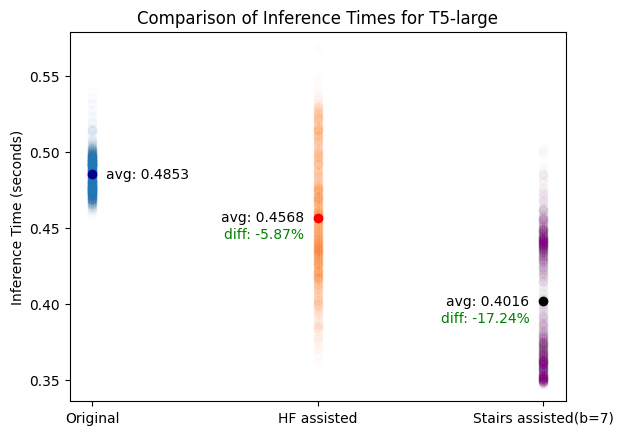}
\label{fig:fig2}
\caption{ Comparison of inference time experiment results for T5-large model. The figure shows a distribution of resulting inference times for the same text generation task with 3 different methodologies -- a standalone model, HugginFace assisted generation, and ``stairs'' assisted generation (300 total runs).}
\end{figure}

Each was run 1000 times after a warmup against the prompt. It can be seen from the results in Figure \ref{fig:fig2} that HuggingFace assisted generation, that is 0.4568 seconds, was 5.87 percent faster than a single T5-model, 0.4853 seconds, with visible distribution of $\pm$0.07 seconds, which in some cases was slower than a plain original model. ``Stairs'' assisted generation on average performed 17.24 percent faster, 0.4016 seconds. Also, it can be seen that it rather consistently performed either 0.04 seconds slower than average or around the same amount faster than average. All generated responses score between 75 and 100 in BLEU score. 

\subsection{T5-3B}

\begin{figure}[hbt!]
\centering
\includegraphics[width=0.5\textwidth]{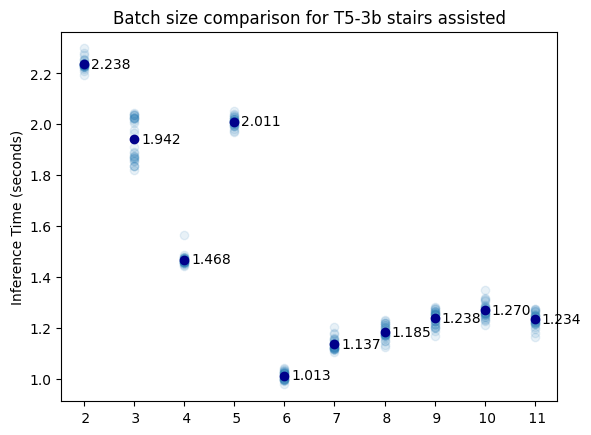}
\caption{\label{fig:fig3}T5-3B stairs assisted generation model inference time in seconds based on batch size (length of assistant model prediction) comparison. The figure displays the distribution of resulting inference times for 10 independent generations of each batch size (100 runs in total).}
\end{figure}

Same as for T5-large experiments, the T5-3B experiment began with batch size selection with the assistance of T5-small-stairs. After running each combination 10 times, same as previously, results indicate that batch size has a noticeable effect on the results. With the best performance being a batch size of 6. All generated responses score between 75 and 100 in BLEU score. This can be seen in Figure \ref{fig:fig3}.

Next, these three cases were compared:
\begin{itemize}
    \item Original -- T5-3B
    \item HF assisted -- HuggingFace T5-3B with assistance of T5-small
    \item Stairs assisted -- T5-3B with assistance of T5-small-stairs with the batch size of 7 (being consistent with T5-large experiments).
\end{itemize}

\begin{figure}[hbt!]
\centering
\includegraphics[width=0.5\textwidth]{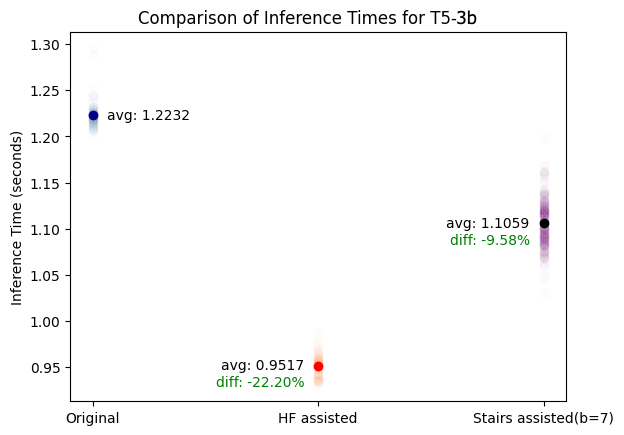}
\label{fig:fig4}
\caption{Comparison of inference time experiment results for T5-3B model. The figure shows a distribution of resulting inference times for the same text generation task with 3 different methodologies -- a standalone model, HugginFace assisted generation, and ``stairs'' assisted generation (30 runs in total).}
\end{figure}

Each was run 100 times after warmup against the prompt. It can be seen from the results in Figure \ref{fig:fig4} that HuggingFace assisted generation, that is 0.9517 seconds, was 22.20 percent faster than a single T5-3B, 1.2232 seconds, with a visible distribution of less than 0.03 seconds. ``Stairs'' assisted generation on average performed 9.58 percent faster, 1.1059 seconds. It was also slightly less consistent than the HuggingFace implementation, mostly varying around $\pm$0.05 seconds. All generated responses score between 75 and 100 in BLEU score.

\section {Conclusions and discussion}\label{conclusions}

The proposed ``stairs'' assisted greedy generation implementation indicates the potential to be a faster inference protocol versus the original model retaining its accuracy. For the T5-large model, the ``stairs'' assisted greedy generation was on average 17.24 percent faster, meanwhile, the HuggingFace implementation was just 5.87 percent faster than the original T5 model. For T5-3B, the proposed  ``stairs'' assisted greedy generation was around 9.58 percent faster, while HuggingFace implementation was around 22.20 percent faster. Additionally, for the proposed implementation, the length of assistant model predictions (batch size) can have up to a 2-time performance increase. Thus, our ``stairs'' assisted greedy generation in specific scenarios has the potential to outperform the production-ready HuggingFace assisted generation.

A proposal to expand these experiments could be using more different prompts: by their lengths, and task variety, or replacing them with more standard evaluation frameworks. The testing environment could be upgraded to better fit T5-3B or similar-size models. On the architectural side, greedy generation could be replaced with sampling, either together for the assistant and main model or individually. That would unlock the temperature parameter. Different sizes and architectures of assistant models or even several different assistant models in parallel could be tested as well.

\bibliography{ref}

\end{document}